\pdfoutput=1
\documentclass[11pt]{article}
\PassOptionsToPackage{hyphens}{url}\usepackage{hyperref}

\usepackage{acl}
\usepackage{times}
\usepackage{latexsym}

\usepackage{times}
\usepackage{latexsym}

\usepackage[T1]{fontenc}
\usepackage[utf8]{inputenc}

\usepackage{microtype}

\usepackage{enumitem}

\usepackage{pdfpages}
\usepackage{subcaption}
\usepackage{booktabs}
\usepackage{siunitx}
\usepackage{amsmath}
\usepackage{graphicx}
\usepackage{tabularx}

\usepackage{xcolor}
\definecolor{blue}{HTML}{058ED9}
\definecolor{red}{HTML}{CC2D35}

\title{Reducing Model Churn: Stable Re-training of Conversational Agents}

\author{Christopher Hidey \qquad  Fei Liu \qquad  Rahul Goel \\
        Google Assistant \\
        \{chrishidey, liufe, goelrahul\}@google.com }
\begin{document}
\maketitle
\begin{abstract}
Retraining modern deep learning systems can lead to variations in model performance even when trained using the same data and hyper-parameters by simply using different random seeds. This phenomenon is known as  \textit{model churn or model jitter}. This issue is often exacerbated in real world settings, where noise may be introduced in the data collection process. In this work we tackle the problem of stable retraining with a novel focus on structured prediction for conversational semantic parsing. We first quantify the model churn  by introducing  metrics for \textit{agreement} between predictions across multiple re-trainings.
%and \textit{exact match-agreement} for the target and all predictions.   
Next, we devise realistic scenarios for noise injection and  demonstrate the effectiveness of various churn reduction techniques such as ensembling and distillation. Lastly, we discuss practical trade-offs between such techniques and show that co-distillation provides a sweet spot in terms of churn reduction with only a modest increase in resource usage.  
% for spoken language understanding (SLU) systems 
%We then show that the churn increases as noise increases and  model size decreases.
%We hope this work provides various practical 

% This document is a supplement to the general instructions for *ACL authors. It contains instructions for using the \LaTeX{} style files for ACL conferences. 
% The document itself conforms to its own specifications, and is therefore an example of what your manuscript should look like.
% These instructions should be used both for papers submitted for review and for final versions of accepted papers.
\end{abstract}

\section{Introduction}

Deep learning systems can perform inconsistently across multiple runs, even when trained on the same data with the same hyper-parameters. Deployment in real-world environments presents a challenge, where constantly changing production systems require frequent re-training of models. For a conversational semantic parsing system such as Google Assistant or Amazon Alexa, where the goal is to convert users' commands into executable forms, this erratic behavior can have some  unfortunate practical consequences. Some examples include irreproducibility, which limits the ability to make meaningful comparisons between experiments \cite{dodge-etal-2019-show, dodge2020finetuning}, bias, which creates credibility issues if systems consistently struggle with members of a certain class \cite{damour2020underspecification}, and user frustration, which can arise due to unpredictable interactions over time. %CH - (cite something?). 

The root cause of widely divergent behavior is underspecification \cite{damour2020underspecification}, where there are many equivalent but distinct solutions to a problem. Non-determinism in model training (e.g. different data orders or weight initializations) can lead to finding local minima that obtain the same measurements on a held-out test set but make different predictions (also known as \emph{model churn}).

\begin{table}
\centering
\footnotesize
\begin{tabular}{p{1.7cm}p{5cm}}

   Query  & will i need snow tires to drive the sierra nevada mountains this afternoon? \\
\midrule
   Model Run 1 & \textcolor{red}{[in:get\_weather [sl:weather\_attribute} \textcolor{blue}{snow tires ] [sl:location sierra mountains ] [sl:date\_time this afternoon ] ]} \\

   Model Run 2 & \textcolor{red}{[in:get\_info\_road\_condition [sl:road\_condition} \textcolor{blue}{snow tires ] [sl:location sierra mountains ] [sl:date\_time this afternoon ] ]}\\

\end{tabular}
\caption{An example from the TOPv2 dataset \cite{ref:TOPv2} where two model runs re-trained on the same data with the same hyper-parameters make \textcolor{red}{different predictions}. Only the first  matches the gold target, but the second has an incorrect intent and slot.}
\label{table:example}
\end{table}

Even in an academic setting, controlling for all non-determinism is unrealistic - Table \ref{table:example} provides an example of churn from the TOPv2 dataset~\cite{ref:TOPv2}. In this case, re-training the same model twice with the same data and hyper-parameters results in two different predictions for the given query. While at the token level the slots and arguments overlap, the  intents are different,  resulting in a drastically different user experience. In this scenario, the dataset is static and yet we still observe model churn.  In a real-world setting, the dataset may be constantly changing and noisy, necessitating frequent re-training. 
The goal, then, is to maintain consistency even in this scenario.

We thus conduct experiments to evaluate and reduce churn across multiple model re-training runs.
%We thus conduct experiments to evaluate and reduce inconsistency across multiple model re-trainings. We experiment with various techniques such as ensembling \cite{Dietterich00ensemblemethods} and distillation/co-distillation \cite{hinton2015distilling, kim-rush-2016-sequence, anil2018large}. We focus specifically on the task of conversational semantic parsing, using the TOP \cite{ref:TOP}, TOPv2 \cite{ref:TOPv2}, MTOP   \cite{li2020mtop}, and SNIPS datasets \cite{coucke2018snips}.
Our contributions are as follows:
\begin{enumerate}
    \item We extend the notion of \emph{model churn} to structured prediction. To this end, we introduce new metrics for \emph{agreement} and \emph{exact match agreement} (Section~\ref{sec:task_def}).
    \item We show that techniques such as ensembling \cite{Dietterich00ensemblemethods} and distillation/co-distillation \cite{hinton2015distilling, kim-rush-2016-sequence, anil2018large}, described in Section~\ref{sec:methods}, reduce churn on the TOP \cite{ref:TOP}, TOPv2 \cite{ref:TOPv2}, MTOP   \cite{li2020mtop}, and SNIPS \cite{coucke2018snips} datasets  (Section~\ref{sec:results}).
    \item We explore the effects of model churn in ``real-world'' environments, conducting experiments with a smaller model and two types of simulated noise (random and systematic)\footnote{Datasets can be found at \url{https://github.com/google/stable-retraining-conversational-agents}} to represent various sources of error (Sections~\ref{sec:expts} and~\ref{sec:results}).
    \item We make practical recommendations based on resource usage (number of parameters) in addition to accuracy and agreement and observe that co-distillation with label smoothing provides the best tradeoff (Section~\ref{sec:discussion}).
\end{enumerate}
To the best of our knowledge, we are the first to study model churn for the structured prediction task of spoken language understanding (SLU). 

%Since metrics like prediction difference~\cite{shamir2020smooth} do not readily extend to structured prediction, we introduce the  metrics \emph{agreement} and \emph{exact match agreement} and show that even on these public academic datasets, there exists substantial disagreement across multiple training runs with the same hyper-parameters (see Table~\ref{table:example} for an example). Then, to imitate real world environments, we conduct experiments with a smaller, more deployable model and add two types of simulated noise (random and systematic) to the data to represent various sources of noise.
%which is easier to deploy

%Further, we conduct  studies to measure the effect of label smoothing, model size, and dataset noise on model churn. We show the effect of various churn-reduction techniques like ensembling and co-distillation. We also do a qualitative error analysis to identify examples with churn that are addressed with churn reduction techniques. Finally, we make practical recommendations based on resource usage (number of parameters) in addition to accuracy and agreement. We observe that co-distillation provides the best trade-off between higher training stability and lower resource usage. To the best of our knowledge, we are the first to study model churn for the structured prediction task of spoken language understanding (SLU). 

\section{Background and Related Work}
The problem of \emph{model churn}~\cite{46326}, defined as the difference in predictions observed across runs when re-training models, has traditionally been studied for classification tasks. In contrast with previous work, we study the problem of model churn for structured prediction, specifically for SLU.
~\citet{shamir2020anti} introduced ``anti-distillation'' to increase diversity in ensemble predictions and ~\citet{shamir2020smooth} introduced the smooth-relu activation function; however, in our initial experiments we did not find significant improvement using these methods when applied to structured prediction.
Other work has explored forms of smoothing to reduce churn, either by computing soft labels using the nearest neighbors~\cite{bahri2021locally} or by weighting the loss term of individual examples using the predicted probabilities from a teacher model~\cite{Jiang2021ChurnRV}. As these methods were developed for classification, we leave the task of adapting them to structured prediction for future work.

Other research has focused on related problems such as reproducibility~\cite{mccoy2019berts} and calibration~\cite{guo2017calibration,mosbach2020stability}. 
~\citet{nie2020can} argue that this phenomenon is due to underlying task complexity and annotator disagreement.
~\citet{damour2020underspecification} claim that reproducibility is primarily due to underspecification, where there are many distinct solutions to the same problem.
While these problems are related to churn, both reproducibility and calibration metrics are computed relative to a target, rather than accounting for agreement across re-training runs. 

It has been well known that ensembling increases reproducibility and model calibration~\cite{hansen1990neural, lakshminarayanan2016simple}. Since ensembles increase inference times, distillation~\cite{hinton2015distilling} is commonly used to train a student model with similar inference resource usage.
~\citet{reich-etal-2020-ensemble} show that ensemble distillation improves calibration for machine translation and named entitity recognition. 
For our distillation baselines, we follow the recipe by ~\citet{chen2019distilling}. For co-distillation, we follow the recipe developed by ~\citet{anil2018large}.
In our work, we look at the aforementioned approaches and compare them in terms of resource usage, churn reduction, and effectiveness on the task of conversational semantic parsing~\cite{ref:TOP,cheng2020conversational, damonte2019practical, ref:SBTOP, lialin2020update}.

\section{Task Definition and Evaluation}
\label{sec:task_def}
We follow recent work \cite{ref:amazon_dont_parse_generate} and treat conversational semantic parsing as sequence generation using auto-regressive neural models. The goal is to make a structured prediction given a user command such as the example in Table~\ref{table:example}.
For structured prediction, the task of \emph{churn reduction} is, given an input, to predict the exact same sequence across multiple re-training runs. A re-training \emph{run} refers to the model parameters that result from different random weight initialization and data order but the same data and hyper-parameters.

Our aim is to reduce churn across runs while maintaining high accuracy on the gold labels. Thus, we report  \textbf{exact match accuracy} (EM) with the mean over $N$ runs. While our goal is not to obtain the state of the art, we do want to show which methods reduce churn without a loss in performance.
%Our aim is for consistent predictions (i.e. lower \emph{model churn}) but with high accuracy on the gold labels. Thus, we report  \textbf{exact match accuracy} (EM) with the mean over $N$ runs. While our goal is not to obtain the state of the art, we do want to show which methods reduce churn without a loss in performance.
 %In making a prediction, we want consistency (i.e. lower \emph{model churn}) but with high accuracy on the gold labels. %In our work there are two primary sub-goals: 1) to match the gold label and 2) to make consistent predictions across model re-training (i.e. lower \emph{model churn}).
%While consistency across re-training is important, no practitioner would choose a reduction in accuracy over an increase in consistency.

To measure churn, we need a way to compare  predictions across runs, independent of the gold labels.
While previous work \cite{shamir2020smooth} has used metrics such as prediction difference (similar to Hamming distance), the focus was on classification tasks only, making it necessary to compute an alternative measure. Metrics such as edit distance or multiple sequence alignment would be appropriate for sequence generation tasks such as machine translation or paraphrasing, where churn across output may differ locally by only a few tokens. Comparatively, the meaning of these metrics is unclear for structured prediction tasks such as semantic parsing. For example, computing a token-level distance between a prediction such as ``[in:unsupported ]'' and ``[in:get\_event [sl:date\_time this weekend ] ]'' would not be a useful measure.
Thus, we report sequence-level model \textbf{agreement} (AGR) across $N$ runs, where each example has a score of 1 if all $N$ runs agree on the exact same predicted sequence and 0 otherwise. However, it is possible for all runs to agree but make an incorrect prediction; the goal ultimately is to consistently make correct predictions. Consequently, we further extend this metric to include the case where the predictions from all $N$ runs agree \emph{and} the predictions match the target.
We refer to this metric as \textbf{exact match agreement} (EM@N).

\section{Methods for Churn Reduction}
\label{sec:methods}

\begin{figure*}[tb]
  \begin{center}
    \begin{subfigure}[b]{\columnwidth}
      \centering
      \includegraphics[width=0.95\columnwidth]{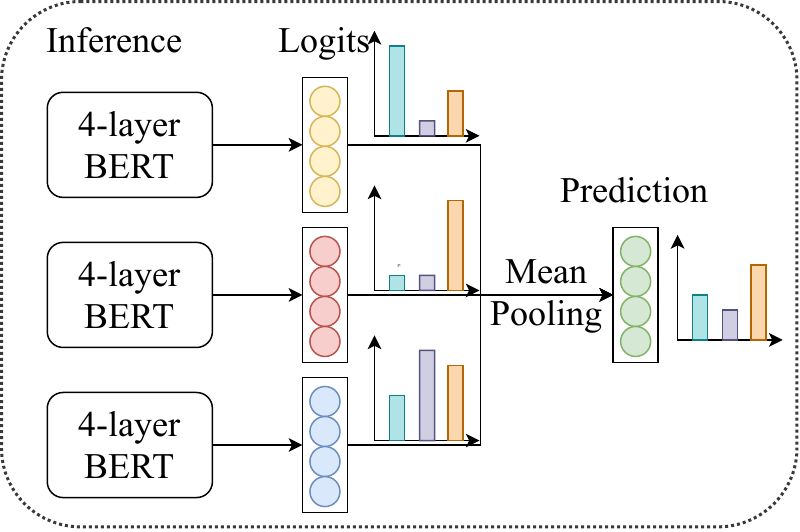}
      \caption{Ensemble.}
    \end{subfigure}
    \begin{subfigure}[b]{\columnwidth}
      \centering
      \includegraphics[width=0.95\columnwidth]{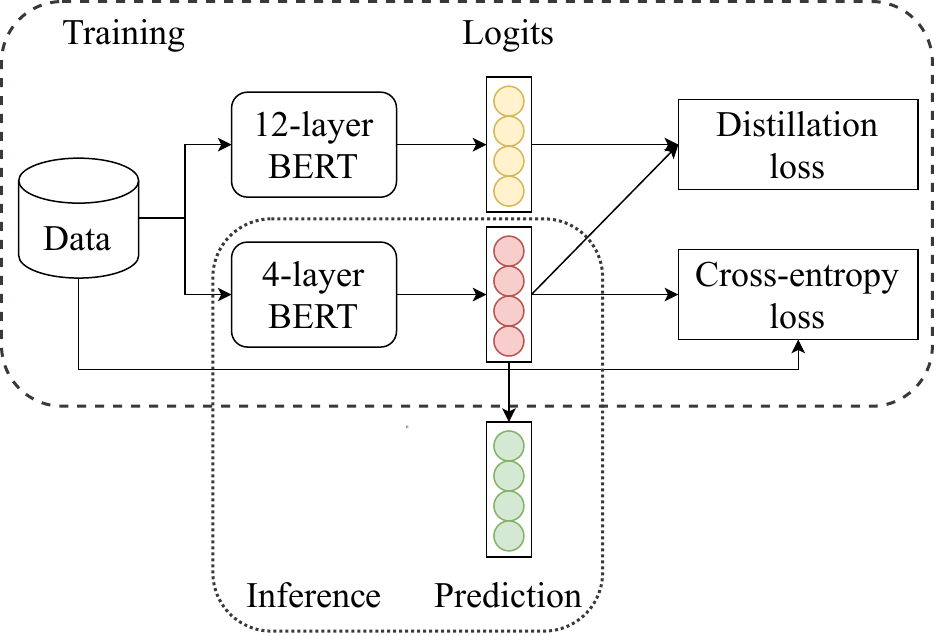}
      \caption{Soft distillation.}
    \end{subfigure}
    \hfill\hfill
    \begin{subfigure}[b]{\columnwidth}
      \centering
      \includegraphics[width=0.95\columnwidth]{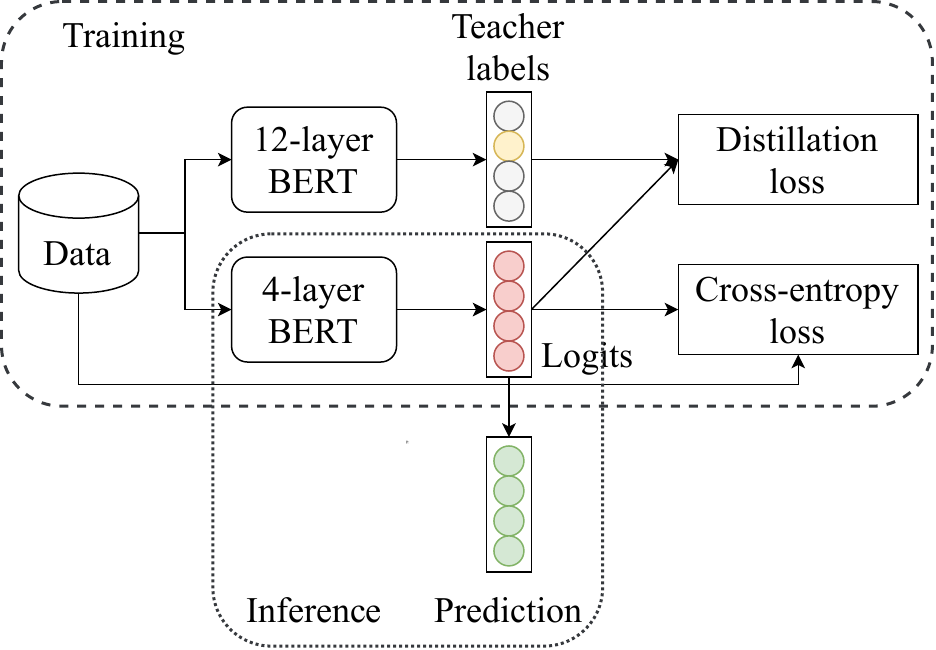}
      \caption{Hard distillation.}
    \end{subfigure}
    \begin{subfigure}[b]{\columnwidth}
      \centering
      \includegraphics[width=0.95\columnwidth]{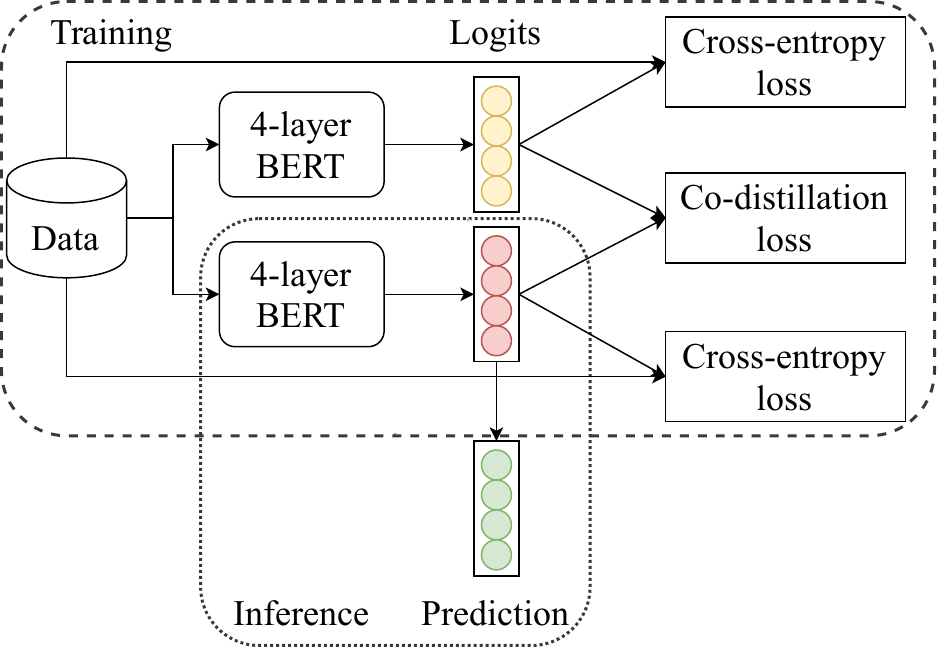}
      \caption{Co-distillation.}
    \end{subfigure}
  \end{center}
  \vspace{-0.25cm}
  \caption{Overview of churn-reducing methods. Dashed and dotted lines indicate the training and inference stages. Rounded rectangular boxes represent seq2seq models with 4- or 12-layer BERT encoders. Ensembling and distillation techniques are applied to the decoder.}
  \label{fig:methods}
\end{figure*}

For our experiments, we explore three  techniques which have been effective on related problems such as model calibration: \textbf{ensembling}, which combines the predictions of multiple models, \textbf{distillation}, which pre-trains a \textit{teacher} model and uses its predictions   to train a \textit{student}, and \textbf{co-distillation}, which trains two or more \textit{peer} models in parallel and allows each model to learn from the predictions of the other. Figure~\ref{fig:methods} displays these techniques.

\subsection{Ensembling}
%A common technique to improve model performance is to train multiple models on the same data and then average their predictions in an ensemble \cite{Dietterich00ensemblemethods}.
We create ensembles by uniformly averaging the probabilities of each model to obtain a point estimate.  As our semantic parser is an auto-regressive sequence-to-sequence model, at every timestep we create the ensemble distribution over the vocabulary from a mixture of $K$ distributions, as in \citet{reich-etal-2020-ensemble}:
\begin{equation}
p(y_t | y_0 ... y_{t-1}, X) = \frac{1}{K} \sum_{k=1}^{K} p_k(y_t | y_0 ... y_{t-1}, X)
\label{eq:ensemble}
\end{equation}
During inference, the next token at each timestep is determined as usual by taking the $argmax$ (in the case of a greedy decoding approach) or using an algorithm such as beam search.  %However, the increased cost of ensembling may be prohibitive in a production system.

\subsection{Distillation}
As ensembling increases model size, distillation \cite{hinton2015distilling} was introduced to compress the knowledge of an ensemble into a single model. With distillation, a \emph{teacher} model\footnote{which is not required to be an ensemble} provides a fixed distribution used to train a \emph{student}. The distillation loss from the teacher can be combined with a loss over the target distribution given by gold labels:
\begin{equation}
\mathcal{L}_{student} =  \mathcal{L}_{NLL}(\theta, \mathcal{D}) + \lambda * \mathcal{L}_{KD}(p_{\theta}, q, \mathcal{D})
\label{eq:distillation_loss}
\end{equation}
where $\mathcal{D}$ is the training dataset, $\mathcal{L}_{NLL}$ is negative log-likelihood loss, and $\mathcal{L}_{KD}$ is knowledge distillation loss. While $\mathcal{L}_{KD}$ may be any dissimilarity measure, we use cross-entropy loss between teacher probabilities $q$ and student probabilities $p_\theta$.

For a sequence generation task, computing the  exact probabilities $q(Y | X)$ and $p(Y | X)$ for a given $X$ is intractable as it would require a computation over the space of all possible $Y$. One way to address this problem is with \emph{sequence-level} distillation \cite{kim-rush-2016-sequence}, which approximates these probabilities with $M$ samples. However, in practice, increasing training time by a factor of $M$ is often infeasible. Instead, we perform \emph{token-level} distillation, computing  token probabilities $q_i$ and $p_i$ at each timestep.

The teacher probability $q_i$ of a token $i$ is computed using the ``softmax'' of its logit $z_i$,\footnote{When distilling from an ensemble, we average the probabilities as in Equation~\ref{eq:ensemble} and convert them back to logits.} adjusted by a \emph{temperature} $T$:
\begin{equation}
    q_i = \frac{exp(z_i/T)}{\sum_j exp(z_j/T)}
\end{equation}
While $T$ usually is set to 1, the temperature can be used to control the entropy of the distribution, where a high temperature increases uniformity.
As the temperature approaches 0, the probability mass is increasingly concentrated on a single token, eventually becoming equivalent to the $argmax$ (a technique known as \textbf{hard distillation}). Otherwise, the method is referred to as \textbf{soft distillation}.

One challenge for distillation is computing the sequence of targets prior to time $t$. One possibility is to perform inference with a method such as beam search to obtain model predictions. Alternatively, we can use teacher-forcing \cite{10.1162/neco.1989.1.2.270, reich-etal-2020-ensemble} and condition on true targets through time $t-1$. For soft distillation, using model predictions would require expensive pre-computation and storage of logits or slower training by performing inference at every timestep. However, for hard distillation, only teacher labels are required, making it possible to pre-compute teacher predictions in a single training set pass.

\subsection{Co-Distillation}
In contrast to distillation, which requires sequential training of the teacher and student, \citet{anil2018large} introduced \textbf{co-distillation}, which involves training multiple \emph{peer} models in parallel. While distillation as an abstract idea only requires logits as a signal, and thus the teacher may be a different architecture or even a different dataset, co-distillation has a few distinct features. First, the peer models share an architecture and training data so that the models can be trained online in parallel. Second, the distillation loss is used before the models have converged. Co-distillation loss is computed as:
\begin{equation}
\begin{split}
        \mathcal{L}_{peers} = & \sum_{k=1}^K  \mathcal{L}_{NLL}(\theta_k, \mathcal{D}) + \\ & \sum_{j \neq k} \lambda * \mathcal{L}_{KD}(p_{\theta_k}, q_j, \mathcal{D})
\end{split}
\label{eq:codistillation_loss}
\end{equation}
where each of $K$ models is trained with negative log likelihood loss ($\mathcal{L}_{NLL}$) on training data as well as  distillation loss ($\mathcal{L}_{KD}$) on the predictions of all other models.

The main advantage of co-distillation is that inference time is equivalent to a single model as only one of the peers is needed. Training time and memory usage are implementation and resource dependent; however the worst case is a K-times increase and may be reduced by, e.g. model parallelism or asynchronous updates \cite{anil2018large}.

\section{Experiment Setup}
\label{sec:expts}
\subsection{Datasets}
\label{subsec:datasets}
We showcase the problem of model churn on 4 conversational semantic parsing datasets. The TOP dataset~\cite{ref:TOP}  consists of queries with hierarchical semantic parses in 2 domains. The TOPv2~\cite{ref:TOPv2} and MTOP~\cite{li2020mtop} datasets expand to 6 more domains with both linear and nested intents and 5 more languages, respectively.\footnote{Although our work is limited to English only.} Table~\ref{table:example} gives an example of the data format shared across all 3 datasets. We further evaluate on SNIPS~\cite{coucke2018snips}, another popular semantic parsing dataset with utterances from 7 domains (including AddToPlaylist, BookRestaurant, GetWeather, and PlayMusic). %(AddToPlaylist, BookRestaurant, GetWeather, PlayMusic, RateBook, SearchCreativeWork, SearchScreeningEvent). 
Data statistics are shown in Table~\ref{tab:datastats}.
%To showcase the problem of model churn we use 4 conversational semantic parsing datasets. The first dataset is TOP~\cite{ref:TOP} which consists of human generated queries and their corresponding hierarchical semantic parses in 2 domains. To show the generalizablity of our results we also use the newer TOPv2~\cite{ref:TOPv2}, which extends the parses from TOP to 6 more domains with both linear and nested intents, and the multilingual MTOP~\cite{li2020mtop}. While TOP and TOPv2 are English focused, MTOP has human annotated parses in 5 more languages.\footnote{We limit our scope to EN only in this work.} Table~\ref{table:example} gives an example of the data format shared across all 3 datasets. Additionally, we further evaluate on SNIPS~\cite{coucke2018snips}, another popular semantic parsing dataset consisting of utterances from 7 domains (AddToPlaylist, BookRestaurant, GetWeather, PlayMusic, RateBook, SearchCreativeWork, SearchScreeningEvent). Data statistics are shown in Table~\ref{tab:datastats}.

\begin{table}[tb!]
\centering
% \resizebox{\textwidth}{!}{
\footnotesize
\begin{tabular}{lrr} 
Dataset & \multicolumn{1}{c}{Train} & \multicolumn{1}{c}{Test} \\
\midrule
TOP & 31,279 & 9,042 \\
TOPv2 & 124,597 & 38,785 \\
MTOP & 15,667 & 4,386 \\ 
SNIPS & 13,784 & 700 \\
\end{tabular}
%}
\caption{Data statistics (\# of utterances).}
\label{tab:datastats}
\end{table}

% \begin{table}[tb!]
% \centering
% % \resizebox{\textwidth}{!}{
% \footnotesize
% \begin{tabular}{lrrr} 
% Dataset & Train & Dev & Test \\
% \midrule
% TOP & 31,279 & 4,462 & 9,042 \\
% TOPv2 & 124,597 & 17,160 & 38,785 \\
% MTOP & 15,667 & 1,772 & 4,386 \\ 
% SNIPS & 13,784 & 0 & 700 \\
% \end{tabular}
% }
% \caption{Data statistics.}
% \label{tab:datastats}
% \end{table}

%Some dataset statistics are shown in Table~\ref{data_stat}. Mention sizes of datasets?
% TOP: 31,279/4,462/9042
% TOPv2: 124,597/17,160/38,785
% MTOP: 
% SNIPS: 13,784/0/700

\subsection{Noise Injection}
\label{subsec:noise}
%We explore the effects of noise when combined with distillation. 
We hypothesize that distillation combined with noise reduces churn without a loss in performance.
On the one hand, adding noise is a common approach to improving model stability and robustness \cite{7780677, NEURIPS2019_f1748d6b}. On the other hand, real-world environments often unintentionally contain noise (due to labels collected from multiple sources, e.g., annotators, users, or distant supervision) and models should be resilient to unexpected changes.
We explore both scenarios, reporting the results of experiments for \textbf{label smoothing} \cite{7780677} for the former and \textbf{random and systematic noise} for the latter. %We conduct these experiments because we hypothesize that model churn increases as noise in the data increases and that distillation can reduce churn without a loss in performance.

\paragraph{Label Smoothing}
Label smoothing is a widely-used technique for calibration of deep learning models, especially for distillation \cite{NEURIPS2019_f1748d6b}. Label smoothing can also be thought of as a noise injection method. This technique is applied by using a weighted average of the one-hot label at a specific timestep and a uniform distribution over all labels.
Specifically, at time step $t$, we compute a new ``soft'' target:
\begin{equation}
    (1 - \alpha) \delta_{t,l} + \alpha \frac{1}{|L|}
\end{equation}
where $\delta_{t,l}$ is the one-hot label if present, $\alpha$ is a parameter that controls the percentage of smoothing, and $L$ is the set of all labels. We follow the recommendations of \cite{NEURIPS2019_f1748d6b} in applying label smoothing only to student models.
We set $\alpha=0.1$ to match the random/systematic noise settings and hold constant the amount of noise across all experiments.

\paragraph{Random Noise}
%To simulate noise that may occur in a real-world scenario, we construct an artificial dataset by randomly swapping X\% of labels from a weighted distribution (refer to Section \ref{subsec:datasets} for details). 
To simulate noise that may occur in a real-world scenario, we create an artificial \textbf{random noise} dataset by randomly swapping 10\% of labels from a weighted distribution.
To construct this dataset, we first find all labels with the prefix ``[in:'' (intents) and compute their probabilities in training. Then, we randomly sample a replacement intent from this distribution. We repeat this process for slots (``[sl:'').

\paragraph{Systematic Noise}
High-quality labeled data for SLU systems may be difficult to come by in large quantities. Conversational agents are therefore often trained using ``distant-labeled'' data from an earlier iteration. This process inevitably results in noisy data, as no SLU system will obtain 100\% on all unseen examples. To simulate this distant supervision, we construct a \textbf{systematic noise} dataset. We train a baseline with a 4-layer BERT encoder (see Section \ref{subsec:impl}) on 90\% of each training set and label the remaining 10\%. However, in order to obtain labels that are both (a) systematic and (b) incorrect, we select the prediction at the \emph{second} beam position rather than the first.\footnote{In practice, this results in less than 10\% of the training data being incorrect. However, on all datasets used in these experiments, the percentage of correct predictions at the second beam position is less than 5\%, thus ensuring that at least 9.5\% of the training data is noisy.}

\subsection{Implementation details}
\label{subsec:impl}
\paragraph{Baselines}
The pointer generator network of \citet{ref:amazon_dont_parse_generate} obtained competitive performance on the TOP datasets using pre-trained encoders. We obtain similar results upon re-implementing this work as a baseline. As our goal is to reduce churn in a realistic environment, we use a  ``production-sized'' encoder -- the 4-layer BERT model of \citet{turc2019wellread} with 4 heads and 256 dimensions -- to reflect what can reasonably be served to users at a robust query-per-second rate. We selected this model to evaluate distillation from a larger model of the same type, 12-layer BERT-base~\cite{ref:bert}, which differs only by the number of parameters. The 4-layer BERT was distilled from BERT-base and obtained only a small  decrease on benchmark datasets compared to larger models. 
%For our pre-trained encoder, we use BERT \cite{ref:bert}, as this model is available in both 12-layer and 4-layer varieties. As our goal is to measure and reduce churn in a realistic environment, we use a ``production-sized model'' -- the 4-layer BERT model of \citet{turc2019wellread} with 4 heads and 256 dimensions -- to reflect what can reasonably be served to users at a robust query-per-second rate. This model is distilled from BERT-base and was shown to obtain only a small  decrease on benchmark datasets compared to larger models.

\begin{table*}[tb!]
\begin{subtable}[tb!]{\textwidth}
\centering
% \resizebox{\textwidth}{!}{
\footnotesize
\begin{tabular}{lcccccccc}
\toprule
& \multicolumn{2}{c}{TOP} & \multicolumn{2}{c}{TOPv2} & \multicolumn{2}{c}{MTOP} & \multicolumn{2}{c}{SNIPS} \\ 
\cmidrule{2-9}
% \cmidrule{3-4}
% \cmidrule{6-7}
% \cmidrule{9-10}
% \cmidrule{12-13}
\multicolumn{1}{l}{Model} & EM (@10) & AGR &  EM (@10) & AGR & EM (@10) & AGR & EM (@10) & AGR  \\
\midrule
BERT-4          &  80.65 (70.29) & 75.48 & 83.88 (73.12) &	78.15 & 79.31 (69.04)	& 73.64 & 86.90 (77.12) & 80.29	\\
Ensemble        & 84.60 (78.55) & 86.18 & 86.42 (80.38) & 88.17 & 84.59 (78.52) & 84.39 & 87.69 (80.58) & 84.60 \\
\midrule 
SD (ensemble) & 81.20 (70.80) &  	76.16 &  	84.00 (73.47) &  	78.75 &  	79.29 	(67.40) &  	71.38 &  	87.29 (79.71) &  	83.45  \\
SD (BERT-12)    &  80.93 (71.14) &  	76.80 &  	84.12 (73.87) &  	79.02 &  	79.23 (68.71) &  	73.23 &  	87.34 (78.27) &  	80.86   			\\
HD (BERT-12)    & 80.72 (70.01) &  	75.03 &  	83.84 (72.57) &  	77.37 &  	78.96 	(68.61) &  	73.07 &  	87.44 \textbf{(80.86)} &  	\textbf{84.75}  \\
Co-distillation       & \textbf{81.43 (73.56)} &  	\textbf{80.41} &  	\textbf{84.21 (76.10)} &  	\textbf{82.99} &  	\textbf{79.45 	(69.73)} &  	\textbf{74.87} &  	\textbf{87.50 (80.86)} &  	\textbf{84.75}  \\
\bottomrule
\end{tabular}
% }
\caption{Original dataset (label smoothing with $\alpha=0.1$).}
\label{tab:0pct}
\end{subtable}
\begin{subtable}[tb!]{\textwidth}
\centering
% \resizebox{\textwidth}{!}{
\footnotesize
\begin{tabular}{lcccccccc}
\toprule
& \multicolumn{2}{c}{TOP} & \multicolumn{2}{c}{TOPv2} & \multicolumn{2}{c}{MTOP} & \multicolumn{2}{c}{SNIPS} \\
\cmidrule{2-9}
% \cmidrule{3-4}
% \cmidrule{6-7}
% \cmidrule{9-10}
% \cmidrule{12-13}
\multicolumn{1}{l}{Model} & EM (@10) & AGR &  EM (@10) & AGR & EM (@10) & AGR & EM (@10) & AGR  \\
\midrule
BERT-4          & 77.02 (65.81) & 71.58 & 82.60 (71.03) & 75.96 & 68.12 (45.88) & 49.12 & 78.41 (57.12) & 58.85 \\
Ensemble      & 78.67 (72.21) & 80.55 & 83.78 (76.53) & 83.89 & 72.37 (58.78) & 65.24 & 82.27 (67.23) & 70.50 \\
\midrule 
SD (ensemble) & 79.44 (68.53) & 73.78 & 83.22 (72.40) & 77.71 & 67.75 (44.51) & 47.23 & 77.89 (56.69) & 58.99 \\
SD (BERT-12)    & 77.11 (65.47) & 71.51 & 82.73 (70.25) & 74.65 & 66.67 (41.00) & 43.62 & 78.11 (56.69) & 58.85 \\
HD (BERT-12)    & 77.33 (59.83) & 63.14 & 82.40 (68.85) & 72.76 & 67.99 (42.84) & 44.51 & 77.89 (56.69) & 58.99 \\
Co-distillation       & \bf{80.21} (\bf{72.04}) & \bf 78.86 & \bf{83.18} (\bf{73.09}) & \bf 78.85 & \bf{73.50} (\bf{58.43}) & \bf 62.22 & \bf{82.00} (\bf{66.33}) & \bf{68.92} \\
\bottomrule
\end{tabular}
% }
\caption{$10\%$ random noise.}
\label{tab:10pct}
\end{subtable}
\begin{subtable}[tb!]{\textwidth}
\centering
% \resizebox{\textwidth}{!}{
\footnotesize
\begin{tabular}{lcccccccc}
\toprule
& \multicolumn{2}{c}{TOP} & \multicolumn{2}{c}{TOPv2} & \multicolumn{2}{c}{MTOP} & \multicolumn{2}{c}{SNIPS} \\
\cmidrule{2-9}
% \cmidrule{3-4}
% \cmidrule{6-7}
% \cmidrule{9-10}
% \cmidrule{12-13}
\multicolumn{1}{l}{Model} & EM (@10) & AGR &  EM (@10) & AGR & EM (@10) & AGR & EM (@10) & AGR  \\
\midrule
BERT-4          & 78.15 (61.36) & 65.11 & 81.80 (67.20) & 70.86 & 74.72 (57.09) & 60.81 & 81.17 (58.42) & 60.43 \\
Ensemble      & 79.87 (68.78) & 74.52 & 83.40 (73.60) & 79.75 & 77.59 (68.55) & 75.80 & 84.50 (71.22) & 74.53 \\
\midrule
SD (ensemble) & 79.85 (67.46) & 72.36 & \bf{83.04} (\bf{71.50}) & \bf 76.60 & 74.84 (57.91) & \bf 61.99 & 81.96 (60.72) & 63.02 \\
SD (BERT-12)    & 79.28 (66.83) & 71.70 & 81.84 (67.47) & 71.10 & 74.97 (57.16) & 61.01 & 81.67 (59.71) & 62.45 \\
HD (BERT-12)    & 79.12 (65.93) & 70.37 & 81.36 (65.33) & 68.47 & 74.51 (56.72) & 60.37 & 80.23 (56.26) & 58.71 \\
Co-distillation       & \bf{80.83} (\bf{72.14}) & \bf 78.45 & 81.97 (70.12) & 75.91 & \bf{75.03} (\bf{58.16}) & 61.49 & \bf{83.66} (\bf{68.78}) & \bf 72.23 \\
\bottomrule
\end{tabular}
% }
\caption{$10\%$ systematic noise.}
\label{tab:10pct_teacher}
\end{subtable}
\caption{Model performance (over $N=10$ runs) when trained on datasets with varying degrees of noise. All student models use 4-layer BERT. BERT-4/12: 4/12-layer BERT. Ensemble: 4-layer ensemble. SD: soft distillation. HD: hard distillation. EM: exact match (mean over $10$ runs). EM@10: EM if all $10$ models are correct. AGR: model agreement. \textbf{Bold}: best non-ensemble.}
\end{table*}

\paragraph{Experiments}
For our experiments, we explore different settings for ensembling and distillation. For both our \textbf{ensemble} and \textbf{ensemble distillation}, we use 4-layer BERT models with $K=3$. We use soft distillation and obtain teacher probabilities  with teacher forcing and Equation \ref{eq:ensemble}. While distilling from an ensemble may increase agreement by preventing the student from assigning too much probability to a single token and becoming overconfident, we also explore \textbf{soft distillation} from a 12-layer teacher. We hypothesize that the 12-layer model would have higher EM but lower AGR than the 4-layer ensemble and this setup allows us to explore any tradeoff between these measurements. 
In addition, we consider \textbf{hard distillation} from a 12-layer model. For this setting, we use beam search inference with a beam width of 3 to obtain predictions, so that we can compare to teacher forcing for soft distillation.
We perform offline inference with the 12-layer model on the entire training set and use both the teacher-labeled data and the gold data for every example.
Finally, we use \textbf{co-distillation} with $K=2$\footnote{as recommended by \citet{anil2018large}.} and $\lambda=1$. We distill from model predictions using weights updated at every timestep. 

\paragraph{Hyperparameters}
To reduce non-determinism, we use a single set of hyper-parameters for the 3 TOP datasets and all experiments. For SNIPS, we select a single set of hyper-parameters by tuning the baseline on 10\% of the training data. Appendix~\ref{appendix:hyper} lists all hyper-parameters.

\section{Results}
\label{sec:results}

We test the effectiveness of the methods described in Section~\ref{sec:methods} over $N=10$ runs. We compile results in Table~\ref{tab:0pct} for models trained on the original datasets with label smoothing. We also report results for the $10\%$ random/systematic noise setting (Tables~\ref{tab:10pct} and \ref{tab:10pct_teacher}) as we assume this represents a ``real-world'' scenario where labels are 90\% correct.

\paragraph{Ensemble superior at the cost of much increased computational cost} First, ensemble sets a high bar in almost all settings regardless of artificial noise. While impressive, this approach requires significantly more computation at inference time and is sometimes deemed infeasible to deploy when accounting for resource usage (see Table~\ref{tab:practical}).

\paragraph{Co-distillation best among distillation-based methods regardless of noise} For label smoothing (Table~\ref{tab:0pct}) and the random/systematic noise settings (Tables~\ref{tab:10pct} and ~\ref{tab:10pct_teacher}), co-distillation clearly and consistently outperforms the baseline in EM, EM@10, and AGR. We also find that soft distillation from the ensemble occasionally obtains the best performance (TOPv2 with systematic noise) but more frequently performs worse than the baseline (MTOP/SNIPS with random noise).  On the other hand, soft/hard distillation perform merely on-par with the baseline or worse. 
Surprisingly, in the $10\%$ random/systematic noise setting, co-distillation not only narrows the gap for EM@10/AGR compared to the ensemble, but also occasionally outperforms the ensemble in EM for TOP/MTOP and TOP, respectively, which may be due to increased robustness to noise during training, rather than only during inference in the ensemble.

\subsection{Effect of Task Difficulty}
Table~\ref{tab:baselines} shows the performance of the baseline models as we increase the task difficulty by reducing the model size or increasing noise in the data. As expected, EM decreases as the task becomes more difficult. However, AGR decreases more rapidly because with lower EM the model has more degrees of freedom to find solutions. These results also show that EM alone is not enough to measure reproducibility and validate the use of EM@10/AGR.

\begin{table}[tb!]
\centering
% \resizebox{\textwidth}{!}{
\footnotesize
\begin{tabular}{lcc} 
\toprule
%& \multicolumn{2}{c}{TOPv2} \\ 
%\cmidrule{2-3}
% \cmidrule{3-4}
% \cmidrule{6-7}
% \cmidrule{9-10}
Model and Setting & EM(@10) & AGR \\
\midrule
%\caption{Going from larger to smaller model increases uncertainity}                                                                                                                                                                                                            \\ \hline
BERT-12 ($0\%$ random noise) & $\textbf{85.68}$ (76.11)  & \textbf{81.30} \\
BERT-4 ($0\%$ random noise) & $83.74$ (73.18)   & 78.15 \\
\midrule
BERT-4 ($10\%$ random noise) &  $82.60$ (71.03)   & 75.96 \\
BERT-4 ($25\%$ random noise) &  $81.34$ (69.04)  & 73.73 \\
BERT-4 ($50\%$ random noise) &  $76.83$ (62.87) & 67.28 \\
\bottomrule
\end{tabular}
%}
%\caption{Baseline model (4/12-layer BERT) performance (over $N=10$ runs) when trained on datasets with varying degrees of noise. EM: exact match accuracy. AGR: model agreement. \textbf{Bold}: best performance.}
\caption{Effect of Task Difficulty on TOPv2, varying baseline model size (4/12-layer BERT) and random noise. EM(@10): exact match (with all 10 runs correct). AGR: model agreement. \textbf{Bold}: best performance.}
\label{tab:baselines}
\end{table}

\subsection{Effect of Label Smoothing}
To better understand the effect of label smoothing, we conduct a study of TOPv2 for the baseline and co-distillation models (Table~\ref{tab:label_smoothing_study})\footnote{see Appendix~\ref{appendix:additional_results} for the full results}. On the base dataset in the baseline setting (BERT-4), label smoothing provides little to no benefit in all metrics. However, we observe a dramatic improvement for co-distillation with label smoothing vs without in EM@10 (+2.14) and AGR (+3.5). On the other hand, on the dataset with 10\% random noise, we do not observe any benefit with label smoothing for either the baseline or co-distillation, perhaps due to the noise already in the data. Finally, on the dataset with 10\% systematic noise, we observe that label smoothing dramatically improves results for both the baseline - EM@10 (+5.07) and AGR (+6.88) - and co-distillation - EM@10 (+2.84) and AGR (+4.59). 
Overall, in the most realistic scenarios (``clean'' or distant-labeled data), we find that co-distillation can be effectively combined with label smoothing. This result is in contrast to \citet{NEURIPS2019_f1748d6b}, who found that training a teacher with label smoothing is not effective. 
When both models are teachers, it is clear that label smoothing helps.

%We present the full results (base dataset with no label smoothing and random/systematic noise with label smoothing) in the Appendix~\ref{appendix:additional_results}.

\begin{table}[tb!]
    \centering
    \footnotesize
    \begin{tabular}{lcc}
    \toprule
    Model and Setting & EM(@10) & AGR \\
    \midrule
    BERT-4 ($\alpha=0$)     &  83.74 (73.18) & 78.47 \\
    BERT-4 ($\alpha=0.1$)     & 83.89 (73.12) & 78.15 \\
    CD ($\alpha=0$)     &  84.01 (73.96) & 79.49 \\
    CD ($\alpha=0.1$)     & \textbf{84.21 (76.10)} &	\textbf{82.99} \\
    \midrule
    BERT-4 ($\alpha=0$, 10\% rand.)     &  82.60 (71.03) & 75.96 \\
    BERT-4 ($\alpha=0.1$, 10\% rand.) & 82.38 (71.11) & 76.24 \\
    CD ($\alpha=0$, 10\% rand.)     &   \textbf{83.18 (73.09)} & 78.85\\
    CD ($\alpha=0.1$, 10\% rand.)     & 82.60 (73.06) & \textbf{79.33} \\
    \midrule
    BERT-4 ($\alpha=0$, 10\% sys.)     &  81.80 (67.20) & 70.86 \\
    BERT-4 ($\alpha=0.1$, 10\% sys.)     & 83.02	(72.27) &	77.74 \\
    CD ($\alpha=0$, 10\% sys.)     &  81.97 (70.12) & 75.91\\
    CD ($\alpha=0.1$, 10\% sys.)     & \textbf{83.19 (73.96)} &	\textbf{80.50} \\
    \bottomrule
\end{tabular}
    \caption{Effects of Label Smoothing on TOPv2. BERT-4: baseline. CD: co-distillation. $\alpha$: label smoothing wt. EM(@10): exact match (with all 10 runs correct) AGR: model agreement \textbf{Bold}: best performance.}
    \label{tab:label_smoothing_study}
\end{table}

\section{Discussion}
\label{sec:discussion}
\paragraph{Qualitative Analysis} To further understand what queries cause the model to churn, we analyze cases where multiple runs disagree. To keep the analysis simple we compare the baseline with co-distillation in Table ~\ref{table:fixed_example} (additional examples in Appendix \ref{appendix:churn_examples}). %using the ensemble as the ceiling. %the best performing co-distillation using the ensemble as the ceiling. 
The first row shows that the baseline model runs are confused by semantically similar slots -- \textit{music\_artist\_name} vs. \textit{music\_track\_title}. The second row demonstrates baseline confusion between the intents \textit{loop\_music} vs. \textit{replay\_music}. In both cases the co-distilled models agree across all training runs. 
%The last 2 rows show that the co-distilled models are confused but the ensembles agree. 
Due to the semantic similarity of the slots/intents, we can attribute this churn to underspecification~\cite{damour2020underspecification}, which is reduced by co-distillation.
% and ensembling  
%leading to variance during inference. 
%As such, methods like ensembling and co-distillation reduce this variance and therefore churn. 

\begin{table}[tb!]
\centering
\footnotesize
\begin{tabular}{p{1.7cm}p{5cm}}

   Query  & play new matchbox 20 \\
\midrule
   Model Run 1 & \textcolor{blue}{{[}in:play\_music {[}sl:music\_artist\_name matchbox 20 {]}{]}} \\

   Model Run 2 & \textcolor{blue}{{[}in:play\_music} \textcolor{red}{{[}sl:music\_track\_title} \textcolor{blue}{matchbox 20 {]}{]}}\\
\midrule
   Query  & repeat closer \\
\midrule
   Model Run 1 & \textcolor{blue}{{[}in:replay\_music {[}sl:music\_track\_title closer {]}{]}} \\

   Model Run 2 & \textcolor{red}{{[}in:loop\_music {]}} \\

\end{tabular}
\caption{Churn examples from TOPv2 fixed by co-distillation. Model predictions are from the baseline. In both cases, only Model Run 1 \textcolor{blue}{matches the target}, but Model Run 2 has an \textcolor{red}{incorrect intent or slot}.}
\label{table:fixed_example}
\end{table}

\begin{figure}[h]
\includegraphics[width=\columnwidth]{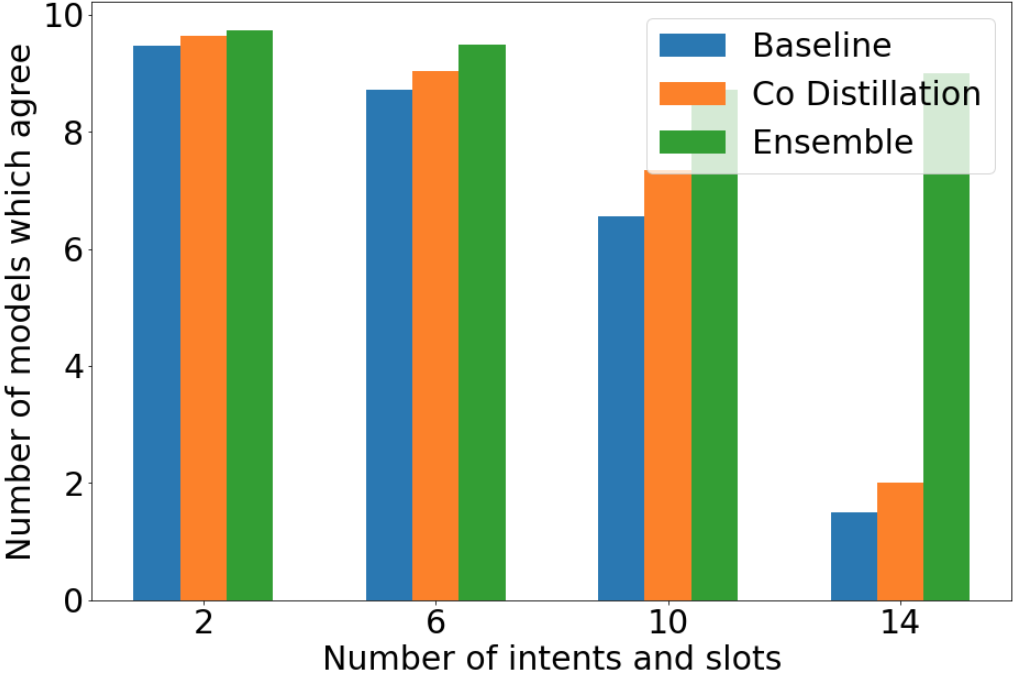}
\caption{Agreement across trained models for various methods vs prediction complexity.}
\label{fig:agreemnt}
\end{figure}

We also explore the relation between agreement and the length of the structured output sequences. Figure~\ref{fig:agreemnt} plots the number of models in agreement against the number of intents and slots.
In making a structured prediction during inference, as length increases the model has more freedom to select incorrect tokens and therefore churn increases.
Co-distillation increases agreement for longer sequences, but ensembling is especially robust. Table~\ref{tab:Average} reports the average target and prediction length where all $N$ models disagree. Surprisingly, we observe that the  models over-generate compared to the target; however, the difference is reduced with co-distillation/ensembling.

%We also ask "Does churn increase as the structured output sequences get longer?". We plot the  of slots and intents vs the number of models which agree in Figure~\ref{fig:agreemnt}. As expected we observe that the model agreement drops as the number of decisions which needs to be made by the model increases. Its at the longer outputs we see a significant difference between performance of baseline, co-distillation and distillation. 

%To inquire into  the predictions when the models disagree, we calculated the average target and prediction length for the queries where all the 10 models don't agree (Table~\ref{tab:Average}). Surprisingly, We observe that the trained models over-generate compared to the target. We also see that the difference gets smaller compared to the baseline for different methods. 

\begin{table}[tb!]
\centering
\footnotesize
%\resizebox{\columnwidth}{!}{%
\begin{tabular}{lcc} %\toprule
% & \multicolumn{2}{c}{Baseline} && \multicolumn{2}{c}{Co-distillation} && \multicolumn{2}{c}{4-layer ensemble} \\ 
%    \cmidrule{2-3} \cmidrule{5-6} \cmidrule{8-9}
Method                 & Target  & Prediction \\ \midrule
Baseline  & 3.66   & 3.91               \\
Co-distillation & 3.77  & 3.82         \\
4 layer ensemble & 3.56  &  3.70 \\
% \bottomrule
\end{tabular}%
%}
%\caption{Average \# of slots and intents for the churn set. We can see that when the model churns, the prediction length is longer than the target. }
\caption{Average \# of slots and intents for cases where all $N$ models disagree. When there is churn the model over-generates (i.e.  prediction length $>$ target length). }
\label{tab:Average}
\end{table}

\paragraph{Practical considerations} We roughly compare the methods along the resource usage dimension in Table~\ref{tab:practical}. As resource usage may be implementation or architecture dependent, we report the number of parameters, which correlates strongly with training/inference time and memory. While ensembling is the strongest approach, it also comes with the most expensive inference. Although wall-clock inference time may be the same as the base model due to parallelization, computing power and memory scales by a factor of $K$. Further, while distillation methods have the same inference time due to similar sized outputs, they have different costs w.r.t. training the teacher.\footnote{Hard/soft distillation have equal number of parameters.} For ensemble distillation, the teacher models can be trained in parallel, but still have $Kx$ storage requirements. For large-model distillation, in practice our 12-layer teacher has about $P=9$ times the number of parameters as the baseline. In both cases, the student \emph{must be trained sequentially}.  Overall, co-distillation performs consistently well across different datasets and noise settings in terms of EM and model agreement while striking a balance between computational cost and performance,
rendering it an attractive approach for  goal-oriented conversational semantic parsing.

\begin{table}[h]
% \resizebox{\columnwidth}{!}{%
\footnotesize
\begin{tabular}{lll}
Method            & Training (actual) & Inference (actual) \\ \midrule
Baseline          & $x$             & $x$             \\
Ensemble    & $P^*_e=3x$             & $P^*_e=3x$             \\
Ens. distillation    & $P^*_e + x = 4x$     & $x$             \\
Large distillation & $P_l + x = 10x$              & $x$  \\
Co-distillation   & $P^*_c=2x $             & $x$ \\
\end{tabular}%
% }
\caption{Overview of resource usage by number of parameters (relative to 4-layer baseline with $x=$\textasciitilde14 million parameters). $P_{e/l/c}$: Number of ensemble/teacher/peer parameters. * denotes parallelism.  }
\label{tab:practical}
\end{table}

% Mention that we measured majority as well. 

% Report flops, should we worry about memory in a handwavy way?

% \usepackage{graphicx}

\section{Conclusion}

Our experiments showed that there exists substantial churn across  runs when re-training models on the same conversational semantic parsing datasets.
We showed that for ``production-sized'' models, co-distillation with label smoothing increases agreement without loss of accuracy. Furthermore, on noisy data simulating a real-world environment, the improvement is even more drastic. When we account for resource usage along with accuracy, we provide strong evidence that co-distillation provides the sweet spot compared to methods like hard/soft distillation and ensembling. 

In future work, we plan to explore how other modeling decisions can increase or decrease model churn. In this work, we limited our focus to BERT encoders with different number of layers. Other questions to explore include whether the choice of pre-training technique affects churn or whether pre-trained encoder-decoders show the same effects. Finally, we will examine whether alternative decoding algorithms, such as non-autoregressive approaches \cite{babu-etal-2021-non, DBLP:journals/corr/abs-2204-06748}, can reduce churn. 

%also adapt classification methods to structured prediction

%When considering resource usage, we find minimal tradeoff for co-distillation as well.

\bibliography{anthology,custom}
\bibliographystyle{acl_natbib}

\appendix

\section{Ethics}
The TOP and SNIPS datasets used in this experiments are intended for research purposes only. We verified that the datasets do not contain personally identifiable information. The risks of dual use for task-oriented conversational semantic parsers are low as we are not performing open-ended generation; however, the models are likely to overfit to certain demographic groups and underperform on others.

\section{Hyper-parameter Search and Settings}
\label{appendix:hyper}

We run our experiments on the TPU v2 available through Google Cloud.\footnote{\url{https://cloud.google.com/tpu}} 

We use the same hyper-parameters for all 3 TOP datasets and SNIPS, except for SNIPS we use a different number of training steps and learning rate. The hyper-parameters were selected using the Google Cloud black box optimizer \cite{46180}. We tuned the parameters using 64 re-runs over the settings described in Table~\ref{table:hparams}. For SNIPS, we held out 10\% of the training data for tuning the training steps (100000) and learning rate ($0.000031$) and trained the final models on 100\% of the training data with the selected hyper-parameters. For distillation experiments we adjusted the learning rate to $1e-5$ and the batch size to 128 to prevent overfitting.

We train all models (including teacher and student) for 300000 steps on the TOP datasets and 100000 on SNIPS.
We use the Adam optimizer with weight decay \cite{DBLP:journals/corr/abs-1711-05101} and the relu activation function. To follow the pointer generator approach of \citet{ref:amazon_dont_parse_generate}, we embed the output vocabulary in 128-dimensional vectors and project the BERT embeddings from the input to 128 dimensions as well. For our transformer decoder \cite{NIPS2017_3f5ee243}, we use 2 heads and 2 layers (see Table~\ref{table:hparams}) with 256 dimensions  for the attention and feed forward layers. 
We also use a maximum output length of 51. We use dropout on the input wordpiece embeddings, after the contextual BERT embeddings, and on the output embeddings before the softmax layer.

\begin{table}[h]
    \centering
    \footnotesize
    \begin{tabular}{|c|c|c|}
    \hline
    Hyper-parameter & Range/Set & Selected Value \\
    \hline
       Learning rate & $[2e-5, 2e-4]$ & 4e-5 \\
       Decoder Heads & $\{2, 4, 8\}$ & 2 \\
       Decoder Layers & $\{2, 4, 8\}$ & 4 \\
       Batch Size & $\{128, 256\}$ & 256 \\
       Dropout & $[0.01, 0.1]$ & 0.0316 \\

    \hline
    \end{tabular}
    \caption{Tuned Hyper-parameters and their Possible Values}
    \label{table:hparams}
\end{table}

\section{Additional Examples}
\label{appendix:churn_examples}

\begin{table*}[h!]
\small
%\resizebox{\textwidth}{!}{%
\begin{tabular}{p{0.2\linewidth}  p{0.33\linewidth}  p{0.0003\linewidth}  p{0.37\linewidth}} \toprule
Query                        & Ground Truth          & & Model predictions   \\ \hline
play new matchbox 20     & {[}in:play\_music {[}sl:music\_artist\_name matchbox 20 {]}{]}  && {[}in:play\_music {[}sl:music\_track\_title matchbox 20 {]}{]}        \\ %\cmidrule{3-4}
    &        && {[}in:play\_music {[}sl:music\_artist\_name matchbox 20 {]}{]}       \\ \hline
repeat closer                & {[}in:replay\_music {[}sl:music\_track\_title closer {]}{]}      && {[}in:replay\_music {[}sl:music\_track\_title closer {]}{]}       \\ %\cmidrule{3-4}
                             &                         & & {[}in:loop\_music {]}         \\ \hline \hline
\multicolumn{4}{c}{Churn examples fixed by co-distillation. Model predictions are from the baseline model}                \\ \hline \hline
show me alarms for tomorrow  & {[}in:get\_alarm {[}sl:date\_time for tomorrow {]}{]} && {[}in:get\_alarm {[}sl:alarm\_name  {[}in:get\_time {[}sl:date\_time for tomorrow {]}{]}{]}{]}  \\ %\cmidrule{3-4}

                             &    && {[}in:get\_alarm {[}sl:date\_time for tomorrow {]}{]}  \\ \hline
                             
take out my wednesday alarm. & {[}in:delete\_alarm {[}sl:alarm\_name && {[}in:delete\_alarm {[}sl:alarm\_name {[}in:get\_time {[}sl:date\_time wednesday {]}{]}{]}{]}  \\ %\cmidrule{3-4}

 &  {[}in:get\_time {[}sl:date\_time wednesday {]}{]}{]}{]}      & & {[}in:silence\_alarm {[}sl:alarm\_name {[}in:get\_time {[}sl:date\_time wednesday {]}{]}{]}{]}  \\   \hline \hline

\multicolumn{4}{c}{Churn examples further fixed by ensembling. Model predictions from the co-distilled model}   \\ \hline \hline

\end{tabular}%
%}
\caption{Qualitative comparison on TOPv2 of the types of errors fixed by co-distillation and ensembling. }
\label{tab:churn_Examples}
\end{table*}

Table~\ref{tab:churn_Examples} provides additional examples where ensembling fixes errors still present in co-distilled models. In these cases, the co-distilled models over-generate (the phenomenon indicated in Table \ref{tab:Average}) whereas the lengths of the ensemble predictions are correctly calibrated to the target lengths.

\section{Additional Results}
\label{appendix:additional_results}
We present the full set of results from Table \ref{tab:label_smoothing_study} in Table \ref{tab:additional_results}. The results in Table \ref{tab:0pct_0alpha} provide strong evidence that co-distillation with label smoothing (Table \ref{tab:10pct_alpha}) is clearly preferable. When we examine the full set of datasets and methods combined with label smoothing in the random/systematic noise setting, we also see that soft distillation from an ensemble performs well. However, in some cases soft ensemble distillation performs worse than the baseline; swapping occasionally slightly better performance for occasionally much worse performance would not be an acceptable tradeoff in most cases. Co-distillation is more stable in terms of consistently outperforming the baseline. Furthermore, co-distillation requires fewer resources and can be trained in parallel.

\begin{table*}[htb!]
\begin{subtable}[tb!]{\textwidth}
\centering
% \resizebox{\textwidth}{!}{
\footnotesize
\begin{tabular}{lcccccccc}
\toprule
& \multicolumn{2}{c}{TOP} & \multicolumn{2}{c}{TOPv2} & \multicolumn{2}{c}{MTOP} & \multicolumn{2}{c}{SNIPS} \\ 
\cmidrule{2-9}
% \cmidrule{3-4}
% \cmidrule{6-7}
% \cmidrule{9-10}
% \cmidrule{12-13}
\multicolumn{1}{l}{Model} & EM (@10) & AGR &  EM (@10) & AGR & EM (@10) & AGR & EM (@10) & AGR  \\
\midrule
BERT-4          & \bf{81.51} (\bf{72.14}) & 77.85 & 83.74 (73.18) & 78.47 & \bf{80.13} (\bf{68.71}) & 72.54 & 86.83 (75.25) & 78.42 \\
Ensemble        & 84.60 (78.55) & 86.18 & 86.42 (80.38) & 88.17 & 84.59 (78.52) & 84.39 & 87.69 (80.58) & 84.60 \\
\midrule 
SD (ensemble) & 81.36 (71.63) & 77.25 & 83.73 (72.62) & 77.72 & 79.50 (68.11) & 71.97 & 86.80 (75.11) & 77.84 \\
SD (BERT-12)    & 81.31 (71.16) & 76.43 & 83.51 (72.13) & 77.10 & 79.87 (67.36) & 70.76 & 86.37 (73.96) & 76.69 \\
HD (BERT-12)    & 81.33 (70.91) & 75.92 & 83.56 (72.15) & 76.99 & 79.66 (67.09) & 70.39 & 86.93 (77.12) & 80.29 \\
Co-distillation       & 81.31 (72.04) & \bf 77.98 & \bf{84.01} (\bf{73.96}) & \bf 79.49 & 79.55 (68.48) & \bf 72.95 & \bf{87.39} (\bf{79.28}) & \bf 82.59 \\
\bottomrule
\end{tabular}
% }
\caption{Original dataset (no noise)}
\label{tab:0pct_0alpha}
\end{subtable}
\begin{subtable}[tb!]{\textwidth}
\centering
% \resizebox{\textwidth}{!}{
\footnotesize
\begin{tabular}{lcccccccc}
\toprule
& \multicolumn{2}{c}{TOP} & \multicolumn{2}{c}{TOPv2} & \multicolumn{2}{c}{MTOP} & \multicolumn{2}{c}{SNIPS} \\
\cmidrule{2-9}
% \cmidrule{3-4}
% \cmidrule{6-7}
% \cmidrule{9-10}
% \cmidrule{12-13}
\multicolumn{1}{l}{Model} & EM (@10) & AGR &  EM (@10) & AGR & EM (@10) & AGR & EM (@10) & AGR  \\
\midrule
BERT-4    &  77.90  	(64.28) &  	68.76 &  	82.39 (71.11) &  	76.24 &  	70.01 (44.97) &  	46.63 &  	76.59 (51.08) &  	52.95    \\
Ensemble      & 78.67 (72.21) & 80.55 & 83.78 (76.53) & 83.89 & 72.37 (58.78) & 65.24 & 82.27 (67.23) & 70.50 \\
\midrule 
SD (ensemble) & \textbf{80.14 (70.59)} &  	\textbf{76.45} & \textbf{83.51 (74.31)} &  	\textbf{80.46} &  	71.14 (48.89) &  	51.27 &  	80.96 (61.01) &  	63.17 \\
SD (BERT-12)    & 78.71 (66.96) &  	72.05 &  	82.71 (70.75) &  	75.50 &  	69.83 (45.26) &  	47.09 &  	78.59 (53.09) &  	55.54 \\
HD (BERT-12)    & 77.71 (64.77) &  	69.54 &  	81.11 (60.39) &  	63.08 &  	69.63 (44.78) &  	46.52 &  	76.70 (47.34) &  	48.92 \\
Co-distillation &  78.91 (68.42) &  	74.54 &  	82.60 (73.07) &  	79.34 &  \textbf{73.74 (57.64)} &  	\textbf{61.22} &  	\textbf{82.50 (68.35)} &  	\textbf{71.80}      \\
\bottomrule
\end{tabular}
% }
\caption{$10\%$ random noise and label smoothing with $\alpha=0.1$.}
\label{tab:10pct_alpha}
\end{subtable}
\begin{subtable}[tb!]{\textwidth}
\centering
% \resizebox{\textwidth}{!}{
\footnotesize
\begin{tabular}{lcccccccc}
\toprule
& \multicolumn{2}{c}{TOP} & \multicolumn{2}{c}{TOPv2} & \multicolumn{2}{c}{MTOP} & \multicolumn{2}{c}{SNIPS} \\
\cmidrule{2-9}
% \cmidrule{3-4}
% \cmidrule{6-7}
% \cmidrule{9-10}
% \cmidrule{12-13}
\multicolumn{1}{l}{Model} & EM (@10) & AGR &  EM (@10) & AGR & EM (@10) & AGR & EM (@10) & AGR  \\
\midrule
BERT-4          &  79.66 (65.28) &  	70.17 &  	83.03 (72.27) &  	77.74 &  	74.58 (58.50) &  	62.86 &  	84.19 (68.20) &  	70.94\\
Ensemble      & 79.87 (68.78) & 74.52 & 83.40 (73.60) & 79.75 & 77.59 (68.55) & 75.80 & 84.50 (71.22) & 74.53 \\
\midrule
SD (ensemble) &  \textbf{81.02} (71.71) &  	77.87 &  	\textbf{83.85 (74.46)} &  	\textbf{80.68} &  	74.97 (58.87) &  	63.30 &  	82.24 (57.12) &  	59.14\\
SD (BERT-12)    & 80.75 (71.22) &  	77.15 &  	83.25 (73.19) &  	78.97 &  	75.01 (59.28) &  	63.30 &  	82.59 (63.02) &  	65.90 \\
HD (BERT-12)    & 79.49 (64.51) &  	69.10 &  	82.93 (72.27) &  	77.94 &  	75.21 (57.11) &  	60.51 &  	81.57 (59.71) &  	62.88 \\
Co-distillation & 80.84 \textbf{(73.61)} &  	\textbf{81.27} &  	83.19 (73.96) &  	80.50 &  	\textbf{76.98 (63.64)} &  	\textbf{68.09} &  	\textbf{85.49 (72.09)} &  	\textbf{76.26} \\
\bottomrule
\end{tabular}
% }
\caption{$10\%$ systematic noise and label smoothing with $\alpha=0.1$.}
\label{tab:10pct_teacher_alpha}
\end{subtable}
\caption{Model performance (over $N=10$ runs) when trained on datasets with varying degrees of noise. All student models use 4-layer BERT. BERT-4/12: 4/12-layer BERT. Ensemble: 4-layer ensemble. SD: soft distillation. HD: hard distillation. EM: exact match (mean over $10$ runs). EM@10: EM if all $10$ models are correct. AGR: model agreement. \textbf{Bold}: best non-ensemble.}
\label{tab:additional_results}
\end{table*}

\end{document}